\documentclass[letterpaper, 10 pt, conference]{ieeeconf}  

\IEEEoverridecommandlockouts                              

\usepackage{amsmath,amsfonts}
\usepackage{graphics} 
\usepackage{epsfig} 
\usepackage{times} 
\usepackage{algorithmic}
\usepackage{array}    
\usepackage{amssymb}  
\usepackage{textcomp}
\usepackage{stfloats}
\usepackage{url}
\usepackage{cite}
\usepackage{booktabs} 
\usepackage[hidelinks]{hyperref}
\usepackage{graphicx}
\usepackage{bm}
\usepackage[justification=centering]{caption} 
\title{\LARGE \bf
Curriculum Reinforcement Learning for Quadrotor Racing with Random Obstacles
}

\author{Fangyu Sun, Fanxing Li, Yu Hu, Linzuo Zhang, Yueqian Liu, Wenxian Yu*, Danping Zou*
\thanks{This work was supported by the National Key Research and Development Program of China (2022YFB3903801) and the National Science Foundation of China (62073214).}
\thanks{Fangyu Sun, Fanxing Li, Yu Hu, Linzuo Zhang, Wenxian Yu, and Danping Zou are with the School of Automation and Perception, Shanghai Jiao Tong University, China (e-mail: dpzou@sjtu.edu.cn).$^{(*)}$ denotes the corresponding author.}
\thanks{Yueqian Liu is with the Faculty of Aerospace Engineering, Delft University of Technology, The Netherlands.}
 }

\begin{document}
\maketitle
\thispagestyle{empty}
\pagestyle{empty}

\begin{abstract}
Autonomous drone racing has attracted increasing interest as a research topic for exploring the limits of agile flight. However, existing studies primarily focus on obstacle-free racetracks, while the perception and dynamic challenges introduced by obstacles remain underexplored, often resulting in low success rates and limited robustness in real-world flight. To this end, we propose a novel vision-based curriculum reinforcement learning framework for training a robust controller capable of addressing unseen obstacles in drone racing. We combine multi-stage cu
rriculum learning, domain randomization, and a multi-scene updating strategy to address the conflicting challenges of obstacle avoidance and gate traversal. Our end-to-end control policy is implemented as a single network, allowing high-speed flight of quadrotors in environments with variable obstacles. Both hardware-in-the-loop and real-world experiments demonstrate that our method achieves faster lap times and higher success rates than existing approaches, effectively advancing drone racing in obstacle-rich environments. The video and code are available at: \url{https://github.com/SJTU-ViSYS-team/CRL-Drone-Racing}.

\end{abstract}

\section{INTRODUCTION}
Vision-based autonomous drone racing in cluttered environments represents a significant challenge, requiring vehicles to fly as fast as possible while avoiding obstacles. Although the remarkable agility of autonomous quadrotors has been demonstrated in controlled laboratory settings \cite{survey}, achieving collision-free racing in densely cluttered environments remains an unresolved problem. By definition, obstacle-aware racing necessitates that the drone maximizes its speed while avoiding obstacles, pushing the platform's dynamics to its operational limits. However, since the gates are also perceived as obstacles, the dual objectives of gate passing and obstacle avoidance are inherently conflicting, posing a significant challenge for both tasks. These competing demands, combined with environmental variability and model mismatches, can lead to catastrophic failures, highlighting the need for robust and adaptive solutions.

\begin{figure}[htbp]
    \centering
    \includegraphics[width=1.0\linewidth]{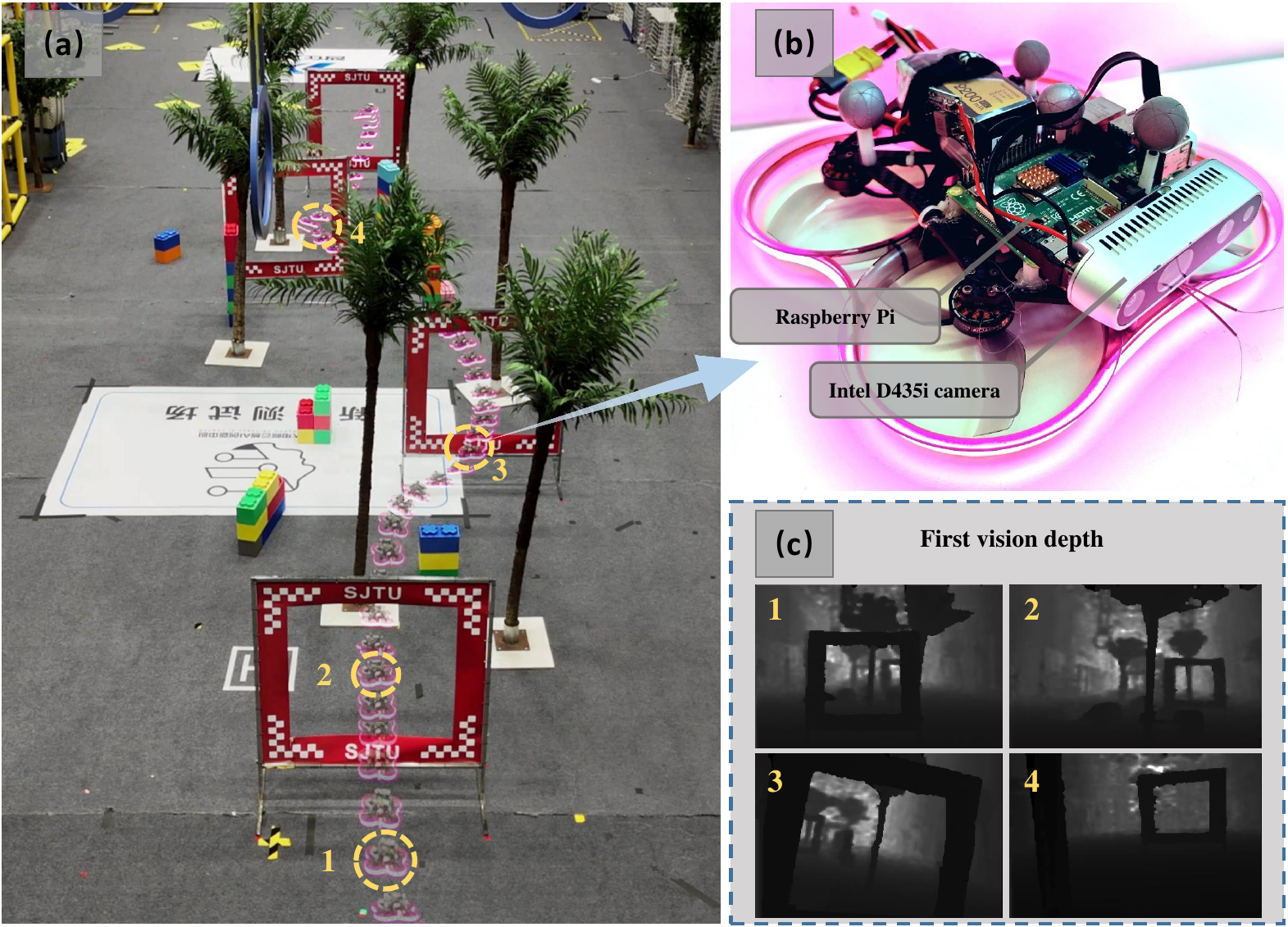}
    \caption{\textbf{Our quadrotor autonomously races with random obstacles in the real world at speeds of up to 8$m/s$.} (a) The real-world experiment of the S-shaped racetrack and the trajectory. (b) Our onboard drone is equipped with a Raspberry Pi computer and an Intel D435i depth camera. (c) FPV depth images during racing with obstacles.}
    \label{fig:image1_ps}
\end{figure}



Recent efforts have sought to integrate obstacle avoidance into drone racing through several methodologies. Traditional path-planning and optimization-based techniques enable high-speed flight and effectively trade off lap time against computational cost \cite{ob_race1,ob_race3,ob_race4}. However, their performance heavily depends on careful algorithm design and can suffer from model mismatches. In contrast, learning-based approaches leverage reinforcement learning (RL) \cite{com,in5} and imitation learning (IL) \cite{ob1} to train policies that achieve low-latency, collision-free control. However, these policies struggle to generalize across racetracks and obstacle configurations, and often fail to transfer reliably from simulation to the real world due to the low success rate.



To address these limitations, we propose a curriculum RL framework for training an end-to-end vision-based control policy, enabling high-speed quadrotor flight in unseen cluttered environments. Our framework employs a multi-stage curriculum learning, allowing the agent to progress from simple obstacle-free navigation to complex racing in cluttered racetracks. 

During training, an obstacle generator with safe margins is incorporated, along with randomized initial states, to facilitate efficient exploration of obstacle-rich environments. To enhance training efficiency, we adapt a customized multi-scene updating mechanism that dynamically adjusts the number of training scenes based on curriculum difficulty, thereby addressing the generalization challenge of policies across diverse obstacle configurations. 

To resolve the inherent conflict between gate traversal and obstacle avoidance, we devise a well-structured reward function that strikes a balance between collision avoidance penalty and gate-passing reward. This reward design also enables rapid yaw maneuvers during high-speed flight, enhancing obstacle awareness while preserving racing performance. Our network architecture is lightweight, which supports real-time onboard inference on a Raspberry Pi. Our main contributions are summarized as follows:
\begin{itemize}
    \item \textbf{Curriculum learning for systematic generalization.} We propose a multi-stage curriculum learning strategy that progressively increases task difficulty through an obstacle generator and random initialization. This addresses the limited generalization of prior methods \cite{com,in5} by enabling the agent to learn robust policies that transfer across unseen cluttered environments.
    \item \textbf{Multi-scene training for improved efficiency.} We introduce a novel multi-scene training paradigm that enhances training efficiency in RL while, more importantly, enabling consistent policy generalization across diverse obstacle configurations.
    \item \textbf{Reward design resolving the conflicting problem.} We design effective reward functions that resolve the inherent conflict between obstacle avoidance and gate passing, allowing the agent to perform high-speed flight with large-angle perception maneuvers without compromising safety.
\end{itemize}

Experimental results demonstrate that our approach consistently outperforms prior methods in both success rate and lap time. Validated through hardware-in-the-loop and real-world flights across three distinct racetracks, our method achieves speeds up to $10\,m/s$ and $8\,m/s$ respectively, with 100\% success rates in all obstacle-rich environments.

\begin{figure*}[htbp]
    \centering
    \includegraphics[width=1.0\linewidth]{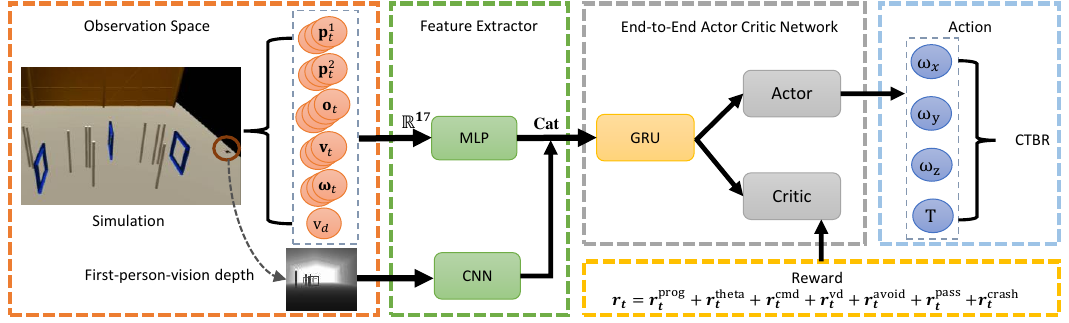}
    \caption{\textbf{The framework of our RL policy training network for the obstacle-rich racing task.} The network architecture consists solely of simple CNNs and MLPs without any complex backbones. "Cat" is the abbreviation for "Concatenate".}
    \label{fig:system1}
\end{figure*}
\section{Related Work}
\subsection{Learning Obstacle-Avoidance Flight}
In recent years, learning-based methods have been increasingly applied to obstacle avoidance for autonomous drones. 
Loquercio et al. \cite{ob_sr} show that an imitation learning (IL) approach can train a neural network entirely in simulation by imitating an expert with full environment knowledge. The resulting policy directly maps noisy onboard sensor observations to collision-free trajectories. Yu et al. \cite{ob2} introduce an RL-based pipeline utilizing depth map inputs, enabling the drone to dynamically adjust its flight speed according to environmental complexity without collision. Similarly, employing RL, Zhao et al. \cite{ob6} propose a hierarchical learning and planning framework that dynamically adjusts the speed constraints of a model-based trajectory planner. Zhang et al. \cite{yuang} leverage the differentiable quadrotor dynamics to learn obstacle-avoidance policy through direct gradient backpropagation. Building on this work, Hu et al. \cite{ob5} replace the depth map with optical flow to achieve obstacle avoidance with a monocular FPV setup. 

However, those methods only consider obstacle avoidance and neglect the requirement of gate traversal. Importantly, these two objectives are inherently conflicting: avoidance encourages conservative maneuvers to steer clear of obstacles, whereas gate traversal demands aggressive and precise flight, often at proximity to the gate center, which itself is perceived as an obstacle.

\subsection{Autonomous Drone Racing}
Autonomous drone racing has emerged as a growing area of research where drones are required to complete a predefined sequence of waypoints in minimal time. Success in this domain necessitates the integration of sophisticated hardware and complex algorithms capable of perceiving the environment, planning optimal paths, and executing actions in real-time. Song et al. \cite{in6} indicate that RL controllers can outperform traditional optimization-based methods, such as those proposed in \cite{oc1} and \cite{oc2}. The emergence of this advantage stems from RL's ability to effectively handle highly nonlinear dynamical systems and complex objectives, which pose significant challenges to traditional optimization methods like model predictive control (MPC) \cite{pampc,mpcc}. Recent research \cite{guanjun} has shown that policies trained with deep RL can even surpass human world champions. Furthermore, recent vision-based drone racing research has demonstrated the ability to bypass traditional state estimation entirely by training an asymmetric actor-critic network to map visual inputs directly to control commands \cite{vi2}.

Despite the strong performance of RL-based methods in obstacle-free racing scenarios, their application to environments with obstacles remains relatively underexplored. To date, only a limited number of studies \cite{com,in5} have addressed this challenging problem. However, these methods either overfit to a single predefined obstacle-aware racing track or underfit due to excessive domain randomization, resulting in low success rates and poor transfer to unseen real-world scenes. Motivated by this gap, we aim to develop a robust policy that adapts to varying obstacle environments while maintaining both high flight speed and success rate. 
\section{Methodology}

\subsection{Dynamics and Task Formulation}
We consider a quadrotor with mass $m$ and diagonal inertia matrix $\mathbf{J}$. Its dynamics are governed by
\begin{align}
    \dot{\mathbf{p}}_W &= \mathbf{v}_W, & \dot{\mathbf{v}}_{W} &= \frac{1}{m}\mathbf{R}_{WB}(\bm f_T + \bm f_D) + \bm{g} \nonumber \\
    \dot{\mathbf{q}} &= \frac{1}{2}\mathbf{q} \hat{\mathbf{\mathbf{\Omega}}}, & \dot{\bm{\omega}} &= \mathbf{J}^{-1}( -\bm{\omega} \times \mathbf{J} \bm{\omega} + \bm\tau_T + \bm\tau_D)
\end{align}
where $\mathbf{p}_W$ and $\mathbf{v}_W$ denote position and velocity in the world frame, $\mathbf{R}_{WB}$ is the rotation matrix from body to world frame, and $\bm{\omega}$ is the angular velocity expressed in the body frame. The term $\hat{\mathbf{\Omega}}$ represents the skew-symmetric matrix of $\bm{\omega}$, while $\bm{g}$ is the gravity vector. The collective thrust along the body $z$-axis and the body torque generated by the four rotors are denoted by $f_T$ and $\bm\tau_T$, respectively. To capture the dynamics of aggressive flight, we incorporate air drag effects through the force and torque terms $\bm f_D$ and $\bm\tau_D$.

The obstacle-rich racing task is framed as an infinite-horizon Markov Decision Process (MDP), characterized by the tuple \((\mathcal{S},\mathcal{A},\mathcal{P},\mathcal{R},\gamma)\), where the state and action spaces $\mathcal{S}$ and $\mathcal{A}$ are continuous. Here, $\mathcal{P}$ denotes the transition probability, $\mathcal{R}$ is the reward function, and $\gamma$ is the discount factor. The objective is to find an optimal policy $\pi^{*}$ that maximizes the expected discounted cumulative reward, formalized as
   \begin{equation}\label{eq:objective}
    \pi^{*}_{\theta} = \operatorname*{argmax}_{\pi} \mathbb{E}_{{\tau} \sim \pi} \left[ \sum_{t=0}^{\infty} \gamma^{t} r(s_t,a_t) \right].
    \end{equation}
In the subsequent sections, we present our vision-based RL controller and the associated policy learning methodology.

\subsection{End-to-End Vision-based Controller}
An overview of our method is given in Fig. \ref{fig:system1}. Our approach directly maps vision-based observations to collective thrust and body rate (CTBR). In the following sections, we introduce key components of our end-to-end controller, including observation and action spaces, reward functions, and the network architecture.

\subsubsection{Observation Space}
The observation space consists of drone's state $\mathbf{s}^{\text{drone}}$ and depth map $\mathbf{s}^{\text{depth}}$. We define the state-based observation as 
\begin{equation}
\mathbf{s}_t^{\text{drone}} = \left[\mathbf{p}_t^{1},\mathbf{p}_t^{2}, \mathbf{v}_t, {v}_d, \mathbf{o}_{t}, \boldsymbol{\omega}_{t} \right] \in \mathbb{R}^{17}.
\end{equation} The $\mathbf{p}_t^{1}\in\mathbb{R}^{3}$ and $\mathbf{p}_t^{2}\in\mathbb{R}^{3}$ correspond to the relative position of the drone to the first and second gate's center. The ${v}_d\in\mathbb{R}^{1}$ is the desired speed, which represents the norm of the velocities along the three axes. The $\mathbf{v}_t\in\mathbb{R}^{3}$, $\mathbf{o}_{t}\in\mathbb{R}^{4}$ and $\boldsymbol{\omega}_{t}\in\mathbb{R}^{3}$ represent the drone's linear velocity, orientation, and angular velocity, respectively. The relative position, linear velocity, and orientation are all transferred to the body axis. Our obstacle-rich racing task also requires additional visual input to guide gate passing and obstacle avoidance. The depth map $\mathbf{s}_t^{\text{depth}} \in\mathbb{R}^{64\times64} $ from a depth camera is used. To reduce the sim-to-real gap caused by vision input, we take the inverse depth and add random Gaussian noise during training. So the whole observation space is $O_t=(\mathbf{s}_t^{\text{drone}},\mathbf{s}_t^{\text{depth}})$.

\subsubsection{Action Space}
The policy is trained to directly map the observation to CTBR commands, defining the action as $\boldsymbol{u}_t=[T,\omega_x,\omega_y,\omega_z] \in \mathbb{R}^{4}$. The first dimension represents the mass-normalized collective thrust, while the last three dimensions represent the angular velocities. 
\begin{figure}[h]
    \centering
    \includegraphics[width=0.43\textwidth]{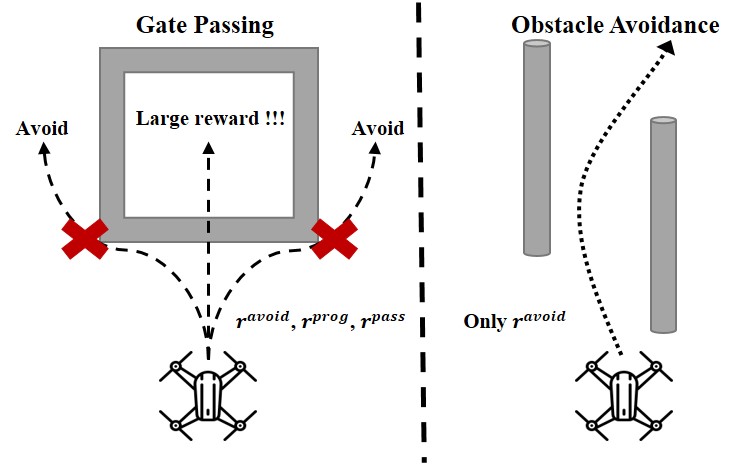}
    \caption{\textbf{Balancing the tasks of gate passing and obstacle avoidance is challenging for training an RL policy.} These two tasks are contradictory for reward design, the obstacle-avoidance reward would cause the drone to bypass the gates from the sides rather than passing through them directly.} %
    \label{fig:avoid1}
\end{figure}
\subsubsection{Reward Functions}
Unlike previous work that focused solely on crossing gates \cite{vi2,rl_race}, our policy needs to introduce additional rewards to avoid obstacles. Moreover, as shown in Fig. \ref{fig:avoid1}, since the gates are also perceived as obstacles, the obstacle-avoidance reward would cause the drone to bypass the gate from the sides when passing through it. Therefore, we need to design reward functions that balance the conflicting tasks of gate passing and obstacle avoidance for the end-to-end control policy. 

We employ a composite reward to guide policy learning for racing in cluttered environments. At each timestep, the overall reward $R_t$ is
\begin{equation}        
R_t=r^{\text{prog}}_t+r^{\text{theta}}_t+r^{\text{cmd}}_t+r^{\text{vd}}_t+r^{\text{avoid}}_t+r^{\text{pass}}_t+r^{\text{crash}}_t,
\end{equation}
where $r_{t}^{\text{prog}}$ encourages progress towards the next gate to be passed \cite{in6}. $r_{t}^{\text{theta}}$ is responsible for aligning the drone to the next gate and for seeing the obstacles. Meanwhile, $r_{t}^{\text{theta}}$ is also the key to enabling the agent to perform large-angle maneuvers based on the gate's pose. $r_{t}^{\text{cmd}}$ is used to penalize large actions, making the trajectory smoother, $r^{\text{vd}}$ penalizes high speed that exceeds the desired velocity. Unlike previous work \cite{com} that relied solely on the discrete reward to achieve obstacle avoidance through trial-and-error during training, we design a simple continuous reward $r_{t}^{\text{avoid}}$ to address the conflict between gate passing and obstacle avoidance tasks. $r_{t}^{\text{pass}}$ and $r_{t}^{\text{crash}}$ are binary rewards that are active only when the drone successfully passes the next gate or a crash happens, respectively. The individual rewards are calculated as follows:
\begin{equation}
    \begin{aligned}
        r_{t}^{\text{prog}} &= \lambda_{1} \left( d_{t-1} - d_{t} \right), \\
        r_{t}^{\text{theta}} &= \lambda_{2}\cdot exp(-\left\| {\theta}_{t}-{\theta}_{\text{gate}}\right\|) ,\\
        r_{t}^{\text{cmd}} &= \lambda_{3} \left\| \boldsymbol{u}_{t} \right\|+\lambda_{4} \left\| \boldsymbol{u}_{t} - \boldsymbol{u}_{t-1} \right\|, \\
        r_{t}^{\text{vd}} &= \lambda_{5}(\left\| \boldsymbol{v}_{t}\right\|-{v}_d) ,\\
        r_{t}^{\text{avoid}} &= \lambda_{6} \left(\frac{1}{d_{col}+b_{\omega}}\right)  ,\\
        r_{t}^{\text{pass}} &= 
        \begin{cases}
        \lambda_7 & \text{if drone passes the next gate} \\
        0 & \text{otherwise}
        \end{cases}\\
        r_{t}^{\text{crash}} &= 
        \begin{cases}
        \lambda_8 & \text{if collision occurs} \\
        0 & \text{otherwise}
        \end{cases}
    \end{aligned}
\end{equation}
where $d_{t}$ represents the distance between the drone and the next gate centers. ${\theta}_{t}$ and ${\theta}_{\text{gate}}$ represent the yaw angle of the drone and the angle from the drone to the next gate's center in the x-y plane, respectively. $\boldsymbol{u}_{t}$ denotes the policy's output. $\boldsymbol{v}_{t}$ represent the linear velocity and ${v}_d$ is the desired speed. $d_{col}$ represents the distance between the drone and the nearest collision point, $b_{\omega}$ is a minor adjustment number. 

\subsubsection{Network Architecture}

As illustrated in Fig. \ref{fig:system1}, the end-to-end architecture comprises a multi-modal feature extractor and an actor-critic network. Specifically, state-based observations and depth maps are encoded by a two-layer MLP and a three-layer CNN, respectively. The resulting feature vectors are concatenated and fed into a Gated Recurrent Unit (GRU) \cite{gru} to capture temporal dependencies. The actor subsequently generates high-level Collective Thrust and Body Rate (CTBR) commands, facilitating direct sim-to-real deployment.
\begin{figure*}[htbp]
    \centering
    \includegraphics[width=1.0\linewidth]{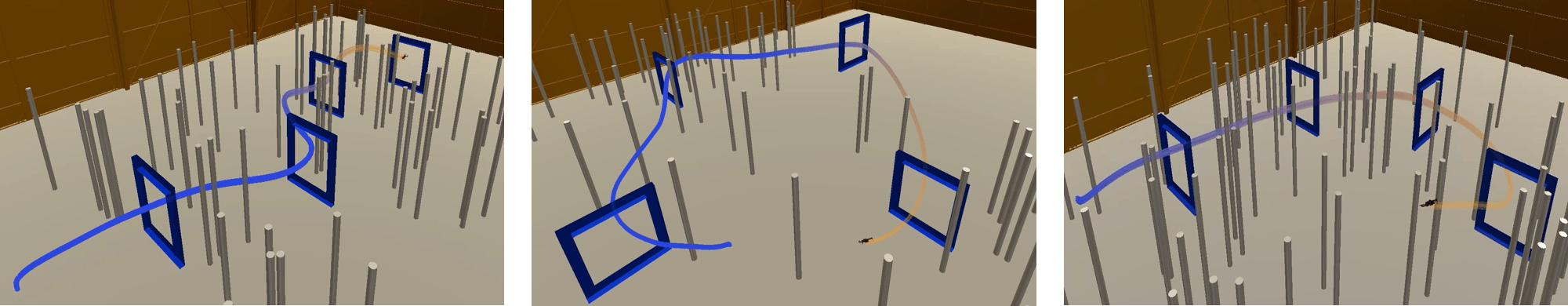}
    \caption{\textbf{Three different racetracks with trajectories flown by our policy in simulation.} It can be observed that under our vision-based policy, the drone can successfully race in densely cluttered environments.}
    \label{fig:demo1-3}
\end{figure*}

\subsection{Policy Training}
We adopt the proximal policy optimization (PPO) algorithm \cite{ppo} for policy training, with the key hyperparameters summarized in Tab. \ref{tab:parameter}. To achieve zero-shot transfer to the real world, the policy must thoroughly explore the observation space and maintain robustness against the sim-to-real gap. We address these requirements by integrating multi-stage curriculum learning with domain randomization techniques.

\subsubsection{Multi-stage Curriculum Learning}
Directly training with a one-step approach causes the RL policy to fall into local optima. Therefore, we propose to design a curriculum to allow the agent to start by completing simple tasks and eventually achieve racing with random obstacles. The designed curriculum has three levels of increasing difficulty to guide the agent's learning process:
\begin{itemize}
  \item Level 1: We train the policy at a lower speed in each racetrack for obstacle-free scenes with vision-based input and whole reward until the agent can finish the racing task.
  \item Level 2: We randomly place obstacles among the waypoints in each racetrack and train the agents to fly at a low desired speed until they can finish the racetrack.
  \item Level 3: Finally, we improve the original desired speed and remove the velocity penalty $r^{\text{vd}}$. We randomize the obstacles within the racetrack and increase their density, training the agents until they have the capability to finish the racetrack at maximum velocity.
\end{itemize}

\subsubsection{Domain Randomization}
Similar to the RL agents presented in previous works \cite{guanjun,rl_race}, we design a domain randomization strategy to robustify a control policy against the sim-to-real gap. First, our strategy needs to account for the stochastic variations of obstacles, enabling the policy to learn a wider range of collision avoidance scenarios. Second, we consider the randomization of gate positions and the initial points of the drone, which is crucial for achieving effective sim-to-real transfer. Therefore, we subdivide our domain randomization approach into two parts: obstacle generator and random initialization strategy.

\textbf{Obstacle Generator} 
We design an obstacle generator to generate random obstacles at each racetrack. Specifically, we regard the area between every two gates as a cuboid and generate a fixed number of obstacles within this space using a random distribution. Our generator ensures that obstacles are always placed on the sections of the racetrack, with a safety distance of $0.5 m$ around each initial point, while guaranteeing the existence of a traversable path through the environment at all times. During training, we reset the environment after a fixed number of time steps to enable the agent to learn from randomized scenarios. Additionally, our generator can specify the number and three-dimensional shapes of each group of obstacles to control the difficulty of the training curriculum. In our simulation, obstacles are modeled as simple geometric primitives with physical collision properties to maintain computational efficiency during training and facilitate reliable sim-to-real transfer.

\textbf{Random Initialization Strategy}
To enhance training efficiency and mitigate the risk of the agent converging to local optima, we ensure that the onboard camera in the simulation captures diverse gate configurations from various initial viewpoints, with all gates within each racetrack being fully visible from these randomized starting points. Concurrently, the initial pose of the drone is uniformly sampled around each starting point, incorporating variations of $\pm0.5m$ in the $x$ and $y$ axes, and $\pm0.5m$ in the $z$ axis. To further improve robustness against uncertainties in gate positioning, we introduced additional randomization to the gate locations, with deviations of $\pm1m$ in the $x$ and $y$ axes, and $\pm0.3m$ in the $z$ axis.

\subsubsection{Multi-scene Updating}
\begin{figure}[htbp]
    \centering
    \includegraphics[width=1.0\linewidth]{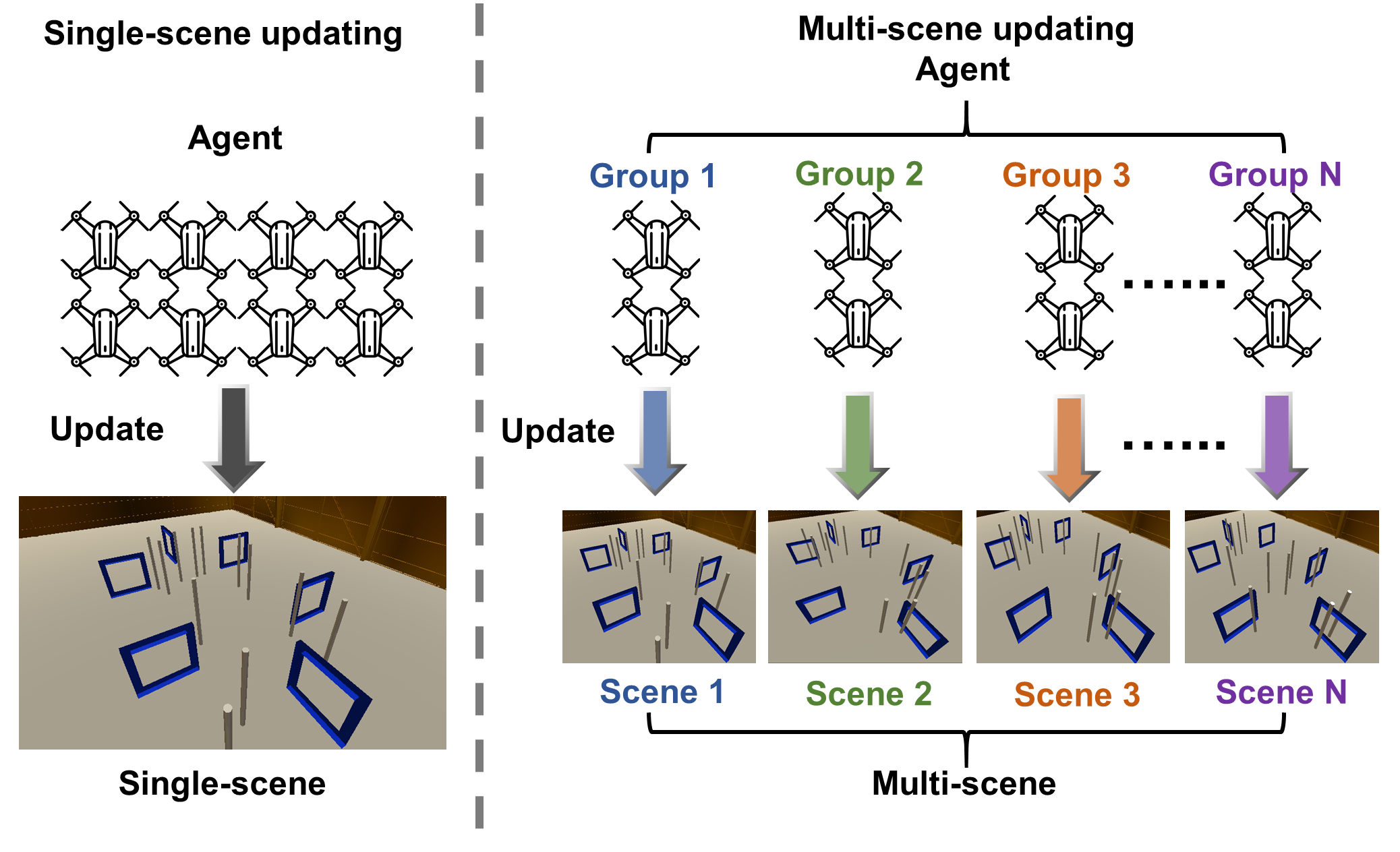}
    \caption{\textbf{The framework of multi-scene updating.} The multi-scene updating approach improves policy training efficiency by customizing the number of scenes in each agent group.}
    \label{fig:multi-scene}
\end{figure}

As shown in Fig. \ref{fig:multi-scene}, we introduce a multi-scene updating technique to significantly enhance RL training efficiency. Traditional RL-based approaches that rely on parallel single-scene updating, where all agents simultaneously update to the same training scene after each rollout \cite{com,in5}. Our method employs a variable number of scenes to update, which supports adjusting the difficulty of curriculum training by customizing the number of parallel-updated scenes, and all agents are trained in groups according to a predefined number of scenes.
This strategy not only improves the robustness of the training process but also accelerates convergence. It is worth mentioning that this method effectively balances the exploration in the early stages of RL with the exploitation in the later stages. It can effectively mitigate policy overfitting, enabling agents to better adapt to unknown environments and achieve superior performance in obstacle-rich racing tasks.

\section{Experiments}
\begin{figure*}[htbp]
    \centering
    \includegraphics[width=0.95\linewidth]{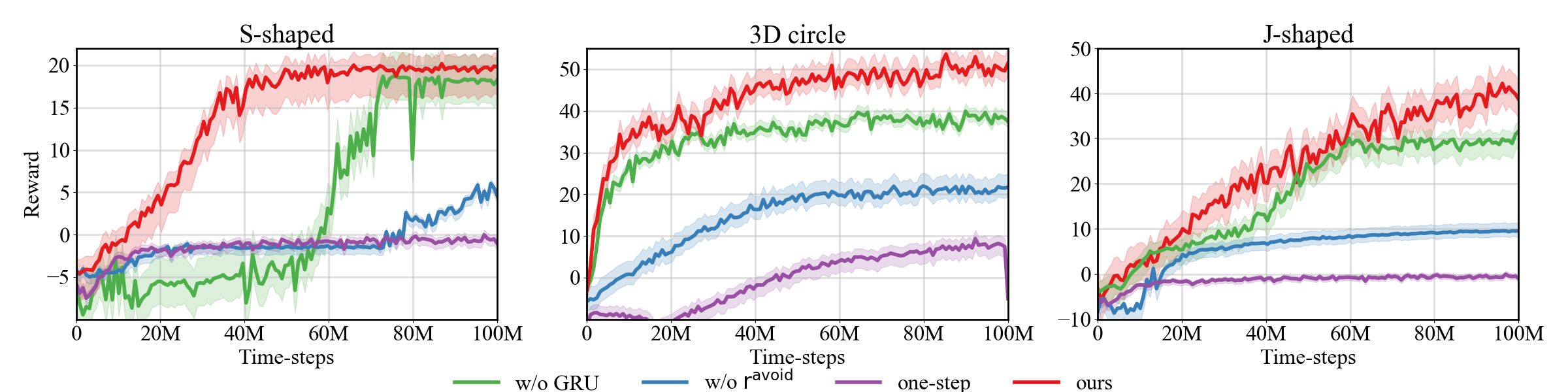}
    \caption{\textbf{The reward curves of the ablation study during training.} Green, blue, purple, and red represent the cases without GRU, without $r_{t}^{\text{avoid}}$, one-step learning, and our method, respectively.}
    \label{fig:ablation}
\end{figure*}
\subsection{Experimental Setup}
We conduct experiments using the VisFly \cite{visfly}, a versatile quadrotor simulator for RL policy training with fast rendering based on Habitat-Sim \cite{Habitat}. We set up three different racetracks (S-shaped, J-shaped, 3D Circle) in obstacle-rich environments. The training is entirely done in simulation, where policy is trained for a total of 100 million time steps in each obstacle-rich racetrack. All the virtual gates and obstacles are equipped with a 3D physics engine and collision detection.


We assess our policy on each racetrack using two metrics: lap time (LT) and success rate (SR). LT measures the racing speed as the total time to complete one full lap, while SR reflects robustness by computing the proportion of successful laps without collision across 10 trials. All experiments run on an Ubuntu 20.04 machine equipped with an i9-13900K processor and an RTX-4090 GPU, achieving a rendering speed of approximately 6000 frames per second for visual inputs.
\begin{table}[h]
\centering
\caption{Key parameters during the experiment.}
\begin{tabular}{lll}
\toprule
 & \textbf{Parameter} & \textbf{Value} \\
\midrule
\textbf{Reward} & $\lambda_1$ & 0.9 \\
 & $\lambda_2$ & 0.05 \\
 & $\lambda_3$  & -0.005 \\
 & $\lambda_4$  & -0.0025 \\
 & $\lambda_5$  & -0.05 \\
 & $\lambda_6$  & -0.01 \\
 & $\lambda_7$ & 5 \\
 & $\lambda_8$ & -4 \\
\midrule
\textbf{PPO} 
 & learning rate & 1e-4 $\rightarrow$ 1e-5 \\
 & discount factor  & 0.99 \\
 & clip range  & 0.2 \\
 & GAE-$\lambda$  & 0.95 \\
 & batch size & 51200 \\
 & number of parallel envs & 100 \\
 & policy network MLP & [192,96] \\
 & value network MLP & [192,96] \\
 & GRU latent dimension & 256 \\
 & number of update scenes & 10 \\
\midrule
\textbf{Quadrotor} & mass [kg] & 0.58 \\
 & inertia [g m$^2$] & [1.01, 1.53, 2.03] \\
 & maximum thrust [N] & 14 \\
 & arm length [m] & 0.075 \\
\bottomrule
\label{tab:parameter}
\end{tabular}
\end{table}

\subsection{Baseline Comparison for Obstacle-aware Racing}

\begin{table*}[htbp!]
\centering
\caption{SR and the LT of our method against two baselines among 10 trials in simulation. Three different racetracks are used to evaluate the performance. The best results are bold.}
\label{tab:performance} 
\renewcommand{\arraystretch}{1.1} 
\setlength{\tabcolsep}{8pt} 
\small 
\begin{tabular*}{\textwidth}{@{\extracolsep{\fill}} c cccccc @{}}
\toprule
\textbf{Method} & \multicolumn{3}{c}{\textbf{LT [s]}} & \multicolumn{3}{c}{\textbf{SR [\%]}} \\
\cmidrule(lr){2-4} \cmidrule(lr){5-7} 
& S-shaped & 3D Circle & J-shaped & S-shaped & 3D Circle & J-shaped \\
\midrule
Vision-based \cite{com} & 5.4 & 3.8 & 3.9 & 30 & 40 & 40 \\
State-based (ours)      & 3.5 & 3.8 & 3.1 & 30 & 20 & 30 \\
\textbf{Vision-based (ours)} & \textbf{3.4} & \textbf{3.6} & \textbf{2.9} & \textbf{100} & \textbf{100} & \textbf{100} \\
\bottomrule
\end{tabular*}
\end{table*}


To evaluate the performance of our policy, we compare it with two baselines: a vision-based policy obtained via RL \cite{com} and a state-based policy trained by our method. The compared vision-based method is also an RL-training policy for obstacle-aware racing, which is only deployed in simulation. In contrast to our method, this approach is to simultaneously perform domain randomization on the racetracks and obstacles during training. Instead of introducing a dedicated reward for obstacle avoidance, it relies solely on discrete penalty terms to learn generalization through trial and error. The state-based policy is our policy training without depth input, including relative positions, velocity, orientation, angle velocity, desired velocity, and the same reward and training techniques. To ensure the fairness of the tests, we employ the same parameter settings of PPO, the same seed, and identical domain randomization methods.

The results are shown in Tab. \ref{tab:performance}, our policy achieves 100\% SR among three different racetracks, with 63\% and 73\% average higher than the other two methods. In terms of LT, our method is comparable to the state-based method, yet it outperforms the vision-based baseline \cite{com}, which is also designed for obstacle-aware racing. Especially in the curvilinear racetracks, our method achieves a significant performance. For the vision-based method \cite{com}, the lack of collision reward during training, combined with the absence of our multi-stage curriculum learning and multi-scene update strategy, results in poor generalization performance and a lower SR. The state-based method, although fast in speed, suffers from low SR due to the lack of visual input guidance for obstacle avoidance. Fig. \ref{fig:demo1-3} demonstrates the corresponding obstacle avoidance trajectories generated by our policy in highly cluttered racetracks.

\subsection{Ablation Study}
We conduct ablative studies to validate the design of our strategy, specifically focusing on three key components: the effectiveness of multi-stage curriculum learning in task completion, the impact of the obstacle-avoidance reward, and the role of GRU. These elements are designed to enable the agent to accomplish the complex task of racing in cluttered environments through progressively increasing difficulty, to learn obstacle avoidance in training, and to process visual-temporal input sequences, respectively.

Tab. \ref{tab:ablation} and Fig. \ref{fig:ablation} present the SR and reward curves of the ablation study, respectively. It can be observed that the performance of our policy deteriorates significantly when any one of the three components is ablated. Among these, replacing the multi-stage curriculum learning with a single one-step training leads to the most severe degradation, causing the policy to converge to a local optimum and fail to accomplish any tasks. Without $r^{\text{avoid}}$, the policy performance also declines notably, with the average SR across all three tracks dropping substantially to merely 36.7\%. Similarly, when the GRU module is removed during training, the lack of temporal information results in slower convergence and reduced success rate. The ablation study validates the importance of the multi-stage curriculum learning, the obstacle avoidance reward, and the GRU module.
\begin{table}[htbp]
\centering
\caption{SR comparison for ablation studies among 10 trials.}
\label{tab:ablation}
\resizebox{\linewidth}{!}{
        \begin{tabular}{ccccc}
        \toprule
         &  & \multicolumn{3}{c}{\textbf{SR [\%]}} \\
        \cmidrule(lr){3-5} 
         &   & S-shaped & 3D Circle & J-shaped \\
        \midrule
        {\textbf{Study. 1}}  
         & w/o GRU  & 80 & 70 & 80 \\
        {\textbf{Study. 2}}  
         & w/o $r^{avoid}$  & 40 & 40 & 30 \\
        {\textbf{Study. 3}}  
         & one-step  & 0 & 0 & 0 \\
        & ours  & \textbf{100} & \textbf{100} & \textbf{100} \\
        \bottomrule
        \end{tabular}
        }
\end{table}

\subsection{Handing RaceTracks and Obstacles Changes}
Our objective is to address the task of racing with random obstacles. Therefore, it is essential to investigate the effectiveness of our method in variable obstacle environments and understand how the performance is influenced by the density of obstacles. Additionally, although this is not our primary focus, we also consider the variability of the racetracks by changing the positions of the gates. Specifically, we conduct two independent sets of evaluations: first, we increase the number of obstacles between every two gates across all tracks from 2 to 5. Second, we separately vary the spatial randomization level of the gates along each axis from $\pm0.3\,m$ to $\pm1.0\,m$.

The results are shown in Tab. \ref{tab:disturbance}. Our policy achieves SR of 76.7\% and 70\% on the most difficult test set, which features maximum obstacle number and gates position changing. The results fully demonstrate that our strategy can robustly handle uncertain obstacle changes, exhibiting good performance even under conditions of dense obstacles and small racetrack adjustments.

\begin{table}[htbp]
\centering
\setlength{\tabcolsep}{3.5pt} 
\caption{SR of handling gates and obstacles across 10 trials. The results correspond to independent evaluations of two factors: (1) Density, which refers to the number of obstacles affecting flight in the racetrack in every 2 gates; and (2) Gate changing (m), which denotes the gate's randomization level among the three axes.}
\label{tab:disturbance}
\resizebox{\linewidth}{!}{
    \begin{tabular}{cccccccc}
    \toprule
     &  & \multicolumn{3}{c}{SR [\%]} \\
    \cmidrule(lr){3-5} 
    & & S-shaped & 3D Circle & J-shaped \\
    \midrule
    {\textbf{Density [its/2 gates]}}
     & 2  & 100 & 100 & 100 \\
     & 3 & 100 & 100 & 100 \\
    & 4 & 90 & 90 & 90 \\
    & 5  & 80 & 70 & 80 \\
    \midrule
    {\textbf{Gate changing [m]}}  
     & $\pm0.3$  & 100 & 100 & 100 \\
     & $\pm0.5$  & 100 & 100 & 100 \\
     & $\pm0.7$  & 90 & 80 & 90 \\
     & $\pm1.0$  & 80 & 60 & 70 \\
    \bottomrule
    \end{tabular}
    }
\end{table}


\subsection{Hardware-in-the-loop Simulation}
\begin{figure*}[htbp]
    \centering
    \includegraphics[width=1.0\linewidth]{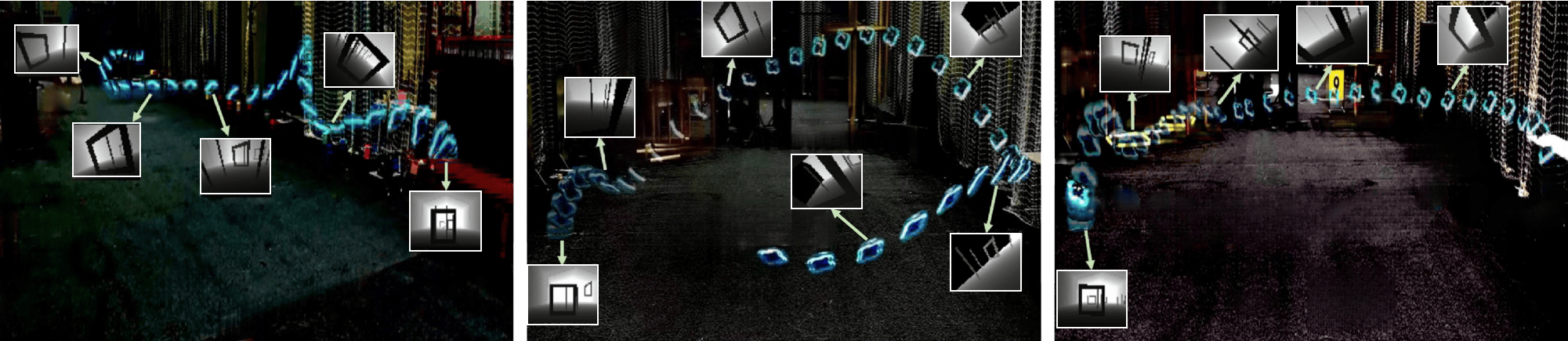}
    \caption{\textbf{HITL simulation.} From left to right, S-shaped, 3D Circle, and J-shaped racetracks.}
    \label{fig:hitl_final}
\end{figure*}

\begin{figure*}[htbp]
    \centering
    \includegraphics[width=1.0\linewidth]{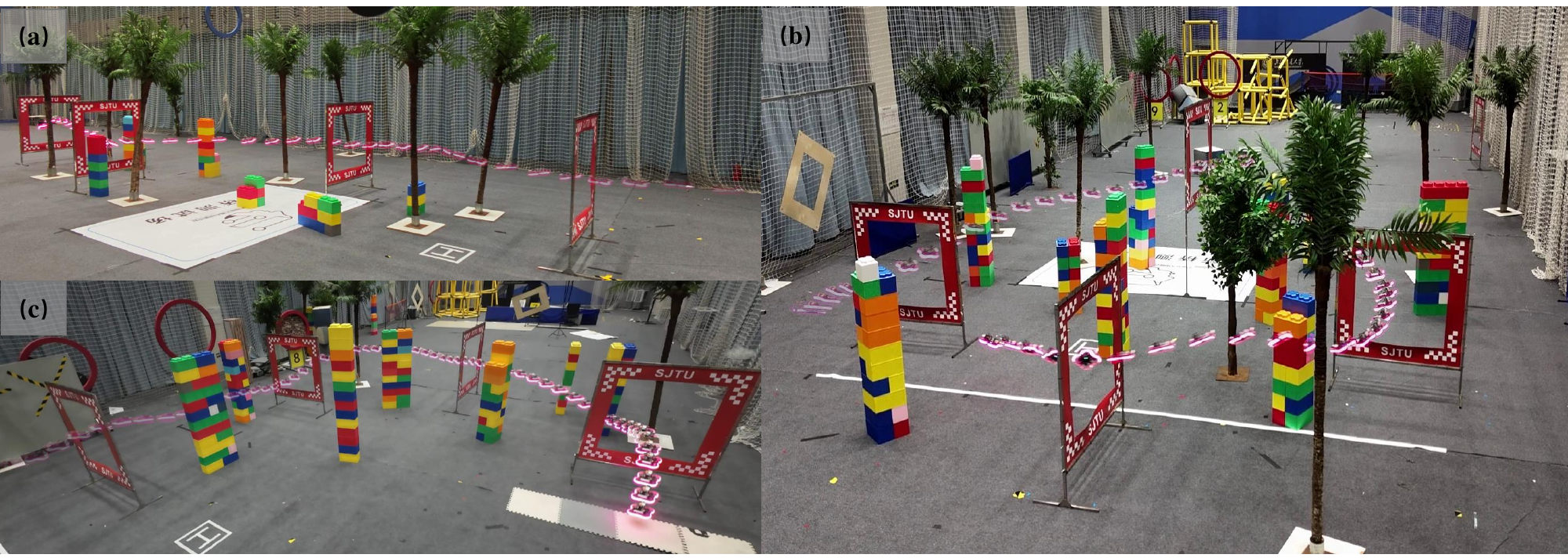}
    \caption{\textbf{Real world experiment results.} We validate our algorithm in three racetracks, including (a) S-shaped, (b) 3D circle, (c) J-shaped.}
    \label{fig:real-world_final}
\end{figure*}
Hardware-in-the-loop (HITL) simulation refers to a method of validating policy using real-world dynamics and visual inputs in simulation, which has been proven to be effective \cite{ob1}. To prevent damage to the onboard computing devices during the initial validation phase, we first employ the dynamics of the real drone and the depth maps corresponding to the positions in the VisFly, ensuring the safety of high-speed flight. Fig. \ref{fig:hitl_final} presents the HITL simulation results along with the first-person-view depth images. It shows that our policy is capable of racing through the gates of each racetrack with visual obstacles at a maximum desired speed $v_d$ exceeding 10$m/s$ while avoiding obstacles along the way. The SR and LT of the HITL simulation are shown in Tab. \ref{tab:hitl_real}.

\begin{table}[h]
\centering
\setlength{\tabcolsep}{3pt} 
\caption{The performance of our policy in HITL simulation and the real world.}
\label{tab:hitl_real}
\resizebox{\linewidth}{!}{
\begin{tabular}{ccccccc}
\toprule
\textbf{Tracks} & \multicolumn{2}{c}{\textbf{S-shaped}} & \multicolumn{2}{c}{\textbf{3D Circle}} & \multicolumn{2}{c}{\textbf{J-shaped}} \\
\cmidrule(lr){2-3} \cmidrule(lr){4-5} \cmidrule(lr){6-7}
& SR [\%] & LT [s]  & SR [\%] & LT [s]  & SR [\%] & LT [s] \\
\midrule
\textbf{HITL} & 100 & 3.5 & 100 & 4.1 & 100 & 3.2 \\
\textbf{Real-world} & 100  & 4.3  & 100 & 5.0  & 100 & 3.7\\
\bottomrule
\end{tabular}
}
\end{table}

\subsection{Real-World Experiment}
Finally, we validate our policy in the real world (see Fig. \ref{fig:image1_ps} for visualization). The entire control pipeline is executed via onboard inference on a Raspberry Pi. Depth data from an Intel D435i camera is downsampled and provided to the policy network at a frequency of 30 Hz. To ensure low-latency performance, the weights are exported in the ONNX \cite{onnx} format, enabling the network to generate control commands at a consistent rate of 30 Hz. These CTBR commands are then processed by the BetaFlight firmware for low-level control. All experiments are conducted in an indoor area using a Vicon motion capture system for state estimation. Each racetrack is evaluated through three separate trials under varying obstacle placements and initial conditions.

The real-world flight performance is shown in Tab. \ref{tab:hitl_real} and Fig. \ref{fig:real-world_final}. For a more dynamic impression, we advise readers to watch the supplementary video showcasing these experiments. Our approach achieves zero-shot sim-to-real transfer of the end-to-end vision-based control policy by aligning real-world dynamics with those observed in simulation. Across three racetracks, the success rate reaches 100\%. Taking into account the obstacle density and the hardware safety constraints of the physical platform, $v_d$ is set to 8 $ m/s$.

\section{Limitations}
In this work, we employ a curriculum RL method for obstacle-aware drone racing, which provides an effective reference for end-to-end control. Nevertheless, due to the inherent exploration and sample-efficiency limitations of RL, our policy can adapt to randomly placed obstacles but fails to generalize to unseen racetrack layouts. This limitation highlights an open problem for current RL-based approaches. In the future, combining the exploratory capacity of RL with the efficiency of differentiable physics learning may enable policies to generalize to random racetracks with obstacles.

\section{CONCLUSIONS}

In this paper, we presented a vision-based curriculum reinforcement learning framework for drone racing in cluttered environments. By integrating a multi-stage curriculum with a multi-scene updating strategy, our approach enables agile flight that balances high-speed gate traversal and reactive obstacle avoidance. Our results demonstrate that the learned policy exhibits significant generalization capabilities across various random obstacle placements. While the current framework assumes a fixed track layout, it effectively accommodates minor perturbations in gate positioning, showing strong robustness during sim-to-real transfer. Future work will focus on extending this generalization to encompass entirely unseen racetrack layouts and more diverse gate geometries, further pushing the limits of autonomous flight in fully unstructured environments.

\addtolength{\textheight}{-12cm}   
\bibliographystyle{IEEEtran} 
\bibliography{Reference.bib}
\end{document}